\documentclass[11pt]{article}

\usepackage{axym-publication}
\usepackage{mathtools}
\usepackage{amsthm}
\usepackage{array}
\usepackage{tabularx}
\usepackage{enumitem}
\usepackage{float}
\usepackage[numbers,sort&compress]{natbib}
\usepackage[capitalise,noabbrev]{cleveref}

\newtheorem{proposition}{Proposition}
\newcolumntype{Y}{>{\raggedright\arraybackslash}X}
\setlist{nosep,leftmargin=1.35em}

\publicationtitle{Temporal Regularized Learning: Self-Supervised Learning
Local in Space and Time}
\publicationauthors{Davide Wiest}
\publicationaffiliations{Department of Computer Science, TU Darmstadt}
\publicationdate{November 2026}
\publicationcode{https://github.com/Axym-Labs/TeReL}
\publicationcorrespondence{dw@axym.org}

\newcommand{\TeReL}{\textsc{TeReL}}
\newcommand{\TeReLOffline}{\TeReL{}-Offline}
\newcommand{\stopgrad}{\operatorname{sg}}
\newcommand{\offdiag}{\operatorname{offdiag}}
\newcommand{\FinalTeReLAccuracy}{96.48}
\newcommand{\FinalTeReLSD}{0.15}
\newcommand{\FinalOfflineAccuracy}{98.41}
\newcommand{\FinalOfflineSD}{0.04}
\newcommand{\FinalSupConAccuracy}{96.95}
\newcommand{\FinalSupConSD}{0.17}
\newcommand{\FinalSupervisedAccuracy}{98.44}
\newcommand{\FinalSupervisedSD}{0.09}

\providecommand{\LeadFigurePath}{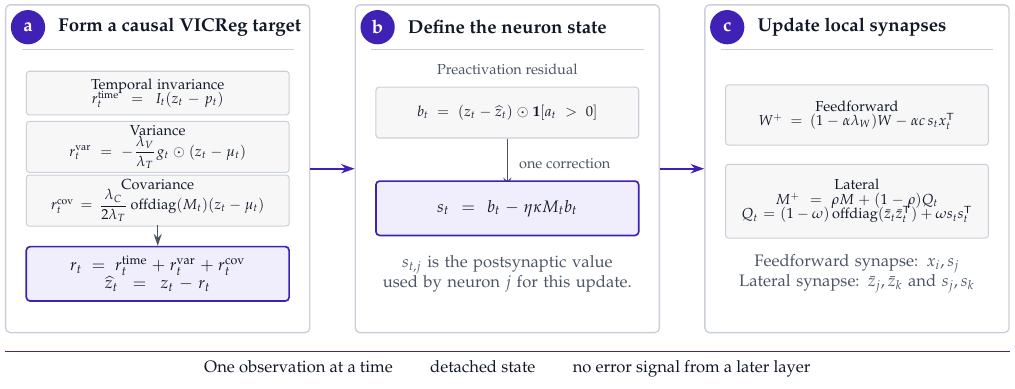}
\providecommand{\LeadFigureCaption}{%
\textbf{One \TeReL{} update in one layer.} \textbf{a}, The temporal, variance,
and covariance terms compare the current activation $z_t$ with the previous
activation $p_t$, the running mean $\mu_t$ and variance gate $g_t$, and the
lateral matrix $M_t$; their sum $r_t$ defines a regularized target
$\widehat z_t$. \textbf{b}, The residual, restricted to active ReLU units and
passed once through the lateral matrix, is the neuron state $s_t$.
\textbf{c}, The state drives a Hebbian feedforward update
($c=2\lambda_T/D$) and an anti-Hebbian lateral update. All stored quantities
are detached, and no error signal passes between layers.}

\begin{document}

\makepublicationtitle

\publicationabstract{%
Self-supervised objectives such as VICReg learn representations by making
related inputs similar, while variance and covariance penalties prevent
collapse. These penalties are statistics over a batch of samples, so
layer-local variants, too, are typically local in space but not in time. We
introduce Temporal Regularized Learning (\TeReL{}), which optimizes a
VICReg-style objective one sample at a time with purely local updates: each
update uses only the current activity of its own and the preceding layer and a
fixed-size running state. Temporally adjacent inputs serve
as positive pairs, which makes the objective a soft form of Slow Feature
Analysis (SFA). Running estimates of each neuron's mean and variance, together
with one lateral matrix per layer, replace the batch statistics; held fixed
during differentiation, they turn the gradient into a Hebbian feedforward
update and an anti-Hebbian lateral update. We show that a
two-layer network trained with batch size one on an MNIST stream of
class-homogeneous chunks reaches $\FinalTeReLAccuracy$\% linear-probe
accuracy, above random features ($95.09$\%) and classical SFA, and that the
temporal order and each loss term contribute to this result. With both
locality constraints relaxed, the same objective matches supervised
backpropagation. \TeReL{} thus shows that the collapse-preventing statistics
of modern self-supervised learning can be held as running state at each
neuron, which brings these objectives closer to what brains and neuromorphic
hardware can implement.}

\makepublicationlinks

\publicationleadfigure[0.26\textheight]{\LeadFigurePath}{\LeadFigureCaption}
\label{fig:overview}

\section{Introduction}

Self-supervised learning trains an encoder without labels by making the
representations of related inputs, called positives, similar. This invariance
objective alone is minimized by a constant representation, so self-supervised
methods add a mechanism that prevents collapse. Contrastive methods push the
representations of unrelated inputs (negatives) apart and typically draw them
from large batches~\cite{chen2020,oord2018}. VICReg and Barlow Twins instead
regularize the statistics of the representation~\cite{bardes2022,zbontar2021}.
VICReg combines three terms~\cite{bardes2022}: an invariance term, the mean
squared distance between the embeddings of two augmented views of an input; a
variance term, a hinge loss that keeps the standard deviation of each
embedding dimension across the batch above a threshold; and a covariance term,
the sum of the squared off-diagonal entries of the batch covariance matrix,
which decorrelates the dimensions.

Both regularizers are statistics over a batch, and the loss is minimized by
backpropagation through all layers. The resulting updates are local neither
in space, because a weight update depends on error signals from later layers,
nor in time, because it depends on the other samples in the batch and on the
activations stored for the backward pass~\cite{rumelhart1986}. Predictive
coding and equilibrium propagation replace the backward pass by iterative
settling over many steps per input~\cite{rao1999,scellier2017}. Layer-local
self-supervised methods remove the backward pass between
layers~\cite{lowe2019,xiong2020,illing2021,siddiqui2024,datta2025}, but most
of them still compute their loss over a batch. Both kinds of locality suit
analog and neuromorphic hardware: spatial locality reduces communication between
units, and temporal locality permits streaming implementations with a small
short-term memory per neuron~\cite{indiveri2015,davies2018}. Classical rules
such as Hebbian learning~\cite{oja1982} and trace
learning~\cite{foldiak1991}, and streaming variants of Slow Feature Analysis
(SFA)~\cite{kompella2012,lipshutz2020}, learn one sample at a time but
optimize simpler objectives. Variance and
covariance regularization, a constraint on a population of samples, appears to
require a batch.

We show that running state at each neuron can take the place of the batch.
Temporal Regularized Learning (\TeReL{}) optimizes a VICReg-style objective one
sample at a time with updates that are local in space and time. It makes three
substitutions, each chosen to keep the update local. First, the positive of an
input is the previous input in the stream rather than a second augmented view:
a stream supplies such pairs without augmentation, and the invariance term
becomes the slowness objective of SFA~\cite{wiskott2002}. Second, each neuron
stores running estimates of its mean and variance in place of the batch
moments, and these estimates are treated as constants during differentiation,
because differentiating through them would require gradients through the
entire history of the stream. Third, one lateral matrix per layer accumulates
running products between pairs of its neurons and supplies a lagged
decorrelation signal in place of the batch covariance. The output of each
layer is detached before it enters the next layer, so no gradient crosses
layers. \Cref{prop:causal-vicreg} shows that these stop-gradients keep a
learning signal from the current input, so the rule needs no batch to learn.

With temporal positives, \TeReL{} is a form of SFA, which extracts slowly
varying features from a sequence of faster-varying
inputs~\cite{wiskott2002,foldiak1991}; where SFA imposes decorrelated,
unit-variance features as a hard whitening constraint, \TeReL{} penalizes
variance and covariance softly, as VICReg does. Slowness is a useful prior
because many sequential domains contain latent factors that evolve more slowly
than their raw observations. In text, the sentiment of a review remains stable
across its many words; in finance, stock prices fluctuate faster than the
market factors that drive them; in physiology, the health condition of a
patient evolves slowly relative to heartbeat, respiratory cycles, and momentary
noise in sensor traces.

Because the loss of each neuron depends only on activations and stored state
within its layer, its gradient takes a familiar form. The gradient at the preactivation of
a neuron is a signed scalar, the neuron state. The feedforward update is the
product of presynaptic activity and this state, and pairs of states drive an
anti-Hebbian update of the lateral matrix. Similarity-matching networks use the
same form of Hebbian cross-layer and anti-Hebbian lateral learning for linear
representations~\cite{pehlevan2015,pehlevan2018}. In \TeReL{}, the state is
instead the residual of a nonlinear target that balances slowness, variance,
and decorrelation.

The experiments test two hypotheses on an MNIST stream ordered in
class-homogeneous chunks. (H1) \TeReL{} trained with batch size one learns a
representation that a linear classifier separates better than random features
of the same architecture and better than classical SFA. (H2) The temporal order
and each of the three loss terms contribute to this representation. The
derivation predicts collapse without the variance term, because the
invariance term alone admits a constant solution, and a shuffled stream
removes the class information that adjacency carries.

\paragraph{Contributions}
\begin{itemize}
  \item We derive \TeReL{} from VICReg (\cref{prop:causal-vicreg}): temporally
  adjacent positives, detached running statistics, and a lagged lateral
  covariance make the loss of each neuron local in space and time while it
  keeps a learning signal from the current input.
  \item We express its gradient as a neuron-local rule: a Hebbian feedforward
  update and an anti-Hebbian lateral update, both driven by one signed neuron
  state, with three state variables per neuron and one $D\times D$ lateral
  matrix per layer. The same derivation casts \TeReL{} as a soft form of SFA
  and relates it to trace learning and similarity matching.
  \item We show that, with batch size one, \TeReL{} reaches
  $\FinalTeReLAccuracy$\% linear-probe accuracy on the MNIST stream, compared
  with $95.09$\% for random features and $90.07$\% for batch SFA; that the same
  objective with relaxed locality matches supervised backpropagation
  ($\FinalOfflineAccuracy$\% versus $\FinalSupervisedAccuracy$\%); and that
  removing the temporal order or any loss term degrades the representation.
\end{itemize}

Running state at each neuron can thus replace the batch statistics that
prevent collapse in modern self-supervised learning, which makes these
objectives available to learning rules that brains and neuromorphic hardware
can implement.

\section{Method}
\label{sec:method}

\subsection{Locality}

We call a learning rule \emph{spatially local} if the update of a weight
depends only on quantities available at its pre- and postsynaptic units (and
same-layer auxiliary units) and does not require backpropagated error signals
from other layers. We call it \emph{temporally local} (online) if the gradient
at time step $t$ depends only on the current observation, the current
parameters, and a bounded amount of detached state carried forward in time.
The size of this state may depend on the layer width but not on the length of
the stream, and no gradient flows through the history that produced it.

\subsection{Starting point: VICReg}

Consider observation $t$ at a layer with input $x_t\in\mathbb R^d$ and width
$D$. The layer computes
\begin{equation}
  a_t=W_tx_t+q_t,
  \qquad z_t=\phi(a_t),
  \label{eq:forward}
\end{equation}
where $W_t\in\mathbb R^{D\times d}$ and $q_t\in\mathbb R^D$. The output $z_t$
is detached before it becomes the input to the next learned layer. Before this
observation, each neuron stores its previous activation $p_t$, a running mean
$\mu_t$, and a running variance $v_t$, and the layer stores a lateral matrix
$M_t\in\mathbb R^{D\times D}$. Let $I_t\in\{0,1\}$ indicate whether $p_t$
belongs to the same sequence, and write $\bar z_t=z_t-\mu_t$ and
$g_t=[\tau-v_t]_+$ for a variance target $\tau>0$.

VICReg decomposes a representation-level objective into three terms
(invariance, variance, covariance)~\cite{bardes2022}:
$\mathcal L_{\mathrm{VICReg}}=\lambda_I\mathcal L_I+
\lambda_V\mathcal L_V+\lambda_C\mathcal L_C$, where $\mathcal L_I$ compares
the embeddings of two views and $\mathcal L_V$ and $\mathcal L_C$ are
functions of the standard deviation and covariance of the embedding
dimensions over a batch. We keep these roles but replace
the second view by the previous activation $p_t$, batch moments by the
detached running statistics $\mu_t,v_t$, and the differentiated batch
covariance by the lagged operator $A_t=\offdiag(M_t)$:
\begin{equation}
\ell_T=\frac{I_t}{D}\lVert z_t-p_t\rVert_2^2,
\qquad
\ell_V=-\frac1D g_t^\top(\bar z_t\odot\bar z_t),
\qquad
\ell_C=\frac1D\bar z_t^\top A_t\stopgrad(\bar z_t).
\label{eq:local-losses}
\end{equation}
The layer minimizes $\ell=\lambda_T\ell_T+\lambda_V\ell_V+\lambda_C\ell_C$
with $\lambda_T>0$. The temporal term makes consecutive activations
similar; in discrete time it is the squared first derivative of the
activations that SFA minimizes. On its own it is minimized by a constant
output, as the invariance term of VICReg is. The variance term rules out this
solution by expanding the coordinates whose stored variance lies below $\tau$,
and the covariance term applies a lagged same-layer decorrelation signal.
\TeReL{} is a causal surrogate of the VICReg decomposition that keeps the
role of each term: $\ell_V$ is a gated instantaneous expansion term, and $A_t$
is a lagged proxy for the covariance gradient. \Cref{sec:results} tests the
contribution of each term.

\begin{proposition}[A stop-gradient VICReg instantiation is temporally local]
\label{prop:causal-vicreg}
For fixed $p_t,\mu_t,v_t,M_t$, and $I_t$, the derivative of $\ell$ is
\begin{equation}
  \nabla_{z_t}\ell=\frac{2\lambda_T}{D}r_t,
  \qquad
  r_t=I_t(z_t-p_t)-\frac{\lambda_V}{\lambda_T}g_t\odot\bar z_t
       +\frac{\lambda_C}{2\lambda_T}A_t\bar z_t,
  \label{eq:causal-vicreg-gradient}
\end{equation}
and the feedforward gradient is $2\lambda_T(J_{\phi,t}^{\top}r_t)x_t^\top/D$,
where $J_{\phi,t}$ is the Jacobian of $\phi$ at $a_t$. The update is
temporally local, and detachment preserves a learning signal from the current
observation whenever $x_t\neq0$ and $J_{\phi,t}^{\top}r_t\neq0$.
\end{proposition}

\begin{proof}
The stored quantities are constants for differentiation. The temporal,
variance, and covariance terms contribute
$2\lambda_T I_t(z_t-p_t)/D$, $-2\lambda_V g_t\odot\bar z_t/D$, and
$\lambda_C A_t\bar z_t/D$, respectively, which sum to
\cref{eq:causal-vicreg-gradient}. All of them are functions of $z_t$ and of
quantities of size $O(D^2)$ computed before observation $t$, so no gradient
reaches an earlier observation. The chain rule through \cref{eq:forward}
gives the outer product $J_{\phi,t}^{\top}\nabla_{z_t}\ell\,x_t^\top$, which is
nonzero when both of its factors are nonzero.
\end{proof}

\subsection{Regularized target and neuron state}

The residual $r_t$ induces an activation target,
$\widehat z_t=\stopgrad(z_t-r_t)$: updates move $z_t$ toward the previous
activation, away from the running mean where the stored variance is below
$\tau$, and against correlated activity. The local loss
$\lambda_T\lVert z_t-\widehat z_t\rVert_2^2/D$ has the same gradient with
respect to $z_t$ as $\ell$. The layer forms this target from its own
activation and state, without an external teacher signal.

Expressed at the input side of the nonlinearity, the residual becomes
$b_t=J_{\phi,t}^{\top}r_t$. For the ReLU encoders used in the experiments,
\begin{equation}
  b_t=r_t\odot\mathbf 1[a_t>0],
  \label{eq:relu-state}
\end{equation}
which requires no automatic-differentiation graph. One pass through the lateral
matrix gives the postsynaptic neuron state
\begin{equation}
  s_t=b_t-\eta\kappa M_tb_t,
  \label{eq:state-correction}
\end{equation}
with correction step $\eta>0$ and strength $\kappa\geq0$. This is a single
matrix--vector product per observation, without iterative settling.

The feedforward update is Hebbian: the product of the input and a
postsynaptic term. With learning rate $\alpha$ and weight decay $\lambda_W$,
\begin{equation}
  W_{t+1}=(1-\alpha\lambda_W)W_t
  -\alpha\frac{2\lambda_T}{D}s_tx_t^\top,
  \label{eq:weight-update}
\end{equation}
which is the gradient step for the detached preactivation target
$\widehat a_t=\stopgrad(a_t-s_t)$. Each feedforward synapse therefore receives
a presynaptic scalar $x_{t,i}$ and a postsynaptic scalar $s_{t,j}$, and under
gradient descent the sign of $s_{t,j}$ determines potentiation or depression.
The bias update uses $s_t$ alone. Without the lateral correction
($\kappa=0$), \cref{eq:weight-update} is exactly the feedforward gradient of
\cref{prop:causal-vicreg}.

\subsection{Lateral matrix and online memory requirements}

The same matrix supplies the covariance signal in
\cref{eq:causal-vicreg-gradient} and the state correction in
\cref{eq:state-correction}. After the synaptic update, it follows
\begin{equation}
  M_{t+1}=\rho M_t+(1-\rho)\big[(1-\omega)\offdiag(\bar z_t\bar z_t^\top)
      +\omega s_ts_t^\top\big],
  \label{eq:lateral-update}
\end{equation}
with $\omega\in[0,1]$ and $\rho\in[0,1)$. With Hebbian-style updates,
mean-centered products between pairs of neurons recover covariance over time.
If the inhibitory lateral weights are written as $L_t=-M_t$, the state
component contributes $-(1-\rho)\omega s_ts_t^\top$ to $L_t$: same-signed
neuron states make their mutual connection more inhibitory, which is the
anti-Hebbian sign.

The \TeReL{} loss requires three per-neuron state variables: the previous
activation, a running mean, and a running variance,
\begin{equation}
  p_{t+1}=z_t,
  \qquad
  \mu_{t+1}=\beta\mu_t+(1-\beta)z_t,
  \qquad
  v_{t+1}=\beta v_t+(1-\beta)(z_t-\mu_t)^2.
  \label{eq:causal-state}
\end{equation}
This state can be stored at the neuron level. The running statistics, the
lateral matrix \cref{eq:lateral-update}, and the feedforward weights
\cref{eq:weight-update} are all leaky integrators (exponential moving
averages), the operation an RC circuit performs in analog hardware. At a
sequence boundary only the validity of $p_t$ is cleared.
A layer of width $D$ thus stores $3D$ dynamic scalars and $D^2$ auxiliary
lateral parameters besides its feedforward weights; plain SGD adds no
optimizer state. For the $784$--$512$--$256$ MNIST encoder these are $2{,}304$
dynamic scalars, $327{,}680$ lateral parameters, and $533{,}248$ feedforward
parameters.

Each layer optimizes its own objective, so \TeReL{} has no global gradient
to approximate and no approximation error that compounds with depth. Each
layer completes its forward pass, target, state correction, and updates for
observation $t$ in one feedforward schedule.

\subsection{\TeReLOffline{}}

\TeReLOffline{} relaxes the temporal and spatial locality constraints within a
subsequence of the data stream. A subsequence enters a ReLU multilayer
perceptron, the same three-term objective is evaluated at its final layer, and
backpropagation differentiates it through all layers and observations of the
subsequence. It isolates the objective from the locality constraints and
is its performance reference.

\section{Related work}
\label{sec:related}

\TeReL{} lies at the intersection of self-supervised learning and local
learning rules. Among self-supervised objectives, contrastive methods prevent
collapse with negative samples~\cite{chen2020,oord2018}; \TeReL{}
requires only temporally adjacent positives and regularizers. VICReg and
Barlow Twins regularize batch statistics of the
representation~\cite{bardes2022,zbontar2021}; \TeReL{} moves the variance and
covariance regularization to running state at each neuron and replaces
augmented pairs by temporally adjacent ones. Many training objectives use
quantities that are computed from parameters or activations but held constant
with respect to the parameters being optimized, such as weight-averaged
teacher parameters~\cite{tarvainen2017}, stop-gradient
targets~\cite{grill2020,chenhe2021}, predictions of an earlier
model~\cite{li2016}, and importance matrices in continual
learning~\cite{kirkpatrick2017,wiest2026iewc}. \TeReL{} uses this pattern for
its running variance and covariance estimates.

Among local learning methods, layer-local and blockwise objectives remove the
backward path between network
modules~\cite{lowe2019,xiong2020,illing2021,siddiqui2024}, and VICReg losses
have been applied at successive layers~\cite{datta2025}. \TeReL{} in addition
makes the update local in time: it supplies the three signals causally along
time and factors them into neuron-local updates. Forward-Forward also trains each
layer with a local objective to avoid end-to-end
backpropagation~\cite{hinton2022}; \TeReL{} uses an invariance objective
instead of a goodness signal and needs no constructed positive and negative
passes. Equilibrium propagation and predictive coding rely on iterative
relaxation, and equilibrium propagation on a free and a nudged
phase~\cite{scellier2017,rao1999,lecun2006}, whereas \TeReL{} uses one
feedforward pass per observation with bounded neuron state.
Hebbian/anti-Hebbian networks derived from similarity-matching objectives have
the same synaptic form as \TeReL{}~\cite{pehlevan2015,pehlevan2018}, and
anti-Hebbian lateral connections decorrelate the units of classical sparse
coding networks~\cite{foldiak1990}. \TeReL{} replaces the linear postsynaptic
factor of similarity matching by the residual of a nonlinear regularized
target.

\paragraph{Relation to slow feature analysis and trace learning}
\TeReL{} is a regularized form of SFA in which temporal adjacency provides
the positive pairs, and it shares the goal of temporal invariance with trace
learning. The squared distance between
two consecutive activations is the squared discrete-time first derivative
that SFA minimizes, and the whitening constraint of SFA corresponds to the
variance and covariance regularizers~\cite{wiskott2002}. Incremental SFA
adapts the linear construction to streams but retains a whitening stage and
linear feature extraction~\cite{kompella2012}. Bio-SFA, a biologically
plausible SFA algorithm, imposes a hard whitening constraint and uses
iterative settling~\cite{lipshutz2020}, whereas \TeReL{} penalizes covariance
softly and trains through a series of discrete-time passes.

Trace learning modifies Hebbian updates by replacing the activation of a
neuron with a temporal trace of it~\cite{foldiak1991,stringer2002}. The two
rules differ in their updates. Ignoring shared factors, the temporal term of \TeReL{} induces
$\Delta w_{kj}\propto x_k(z_{t-1,j}-z_{t,j})$, while trace learning uses
$\Delta w_{kj}\propto x_k(\beta\bar z_{t-1,j}+(1-\beta)z_{t,j})$ with
$\beta\in(0,1)$. No choice of $\beta$ makes the two updates identical, because
matching them would require both $\beta=1$ and $1-\beta=-1$. The trace rule
also requires a weight-dependent decay term to keep the weights
bounded~\cite{foldiak1991}. In \TeReL{}, the ablations show that the variance
and covariance terms are both necessary to avoid collapse or divergence
(\cref{sec:results}), which makes explicit variance control and
decorrelation a candidate addition to trace-learning rules.

\section{Experimental protocol}
\label{sec:experiments}

We evaluate \TeReL{} on MNIST~\cite{lecun1998} presented as a stream. \TeReL{}
uses no labels: its positives are whichever inputs the stream places next to
each other, so the stream order determines what the representation becomes
invariant to. We order the stream so that adjacency carries class
information: it consists of randomly permuted class-homogeneous chunks of 16
samples, separated by sequence boundaries. The individual images are then the
fast-varying inputs, and the class shared within a chunk is the slow feature.
The labels determine adjacency but enter neither the target nor any update. As
Hinton notes for the supervised Forward-Forward setup, the network should then
preferentially learn features that discriminate between
classes~\cite{hinton2022}.

The encoder is a fully connected ReLU network with hidden widths 512 and 256.
As in the Forward-Forward experiments~\cite{hinton2022}, the networks use no
weight sharing: a shared weight would receive the sum of updates from many
neurons, so its update would no longer be local to one synapse. \TeReL{} is
trained with batch size one for two passes over the encoder-training samples.
A linear classifier, trained on the standardized concatenated activations of
both layers for the encoder-training samples, measures representation quality
on held-out data. A stratified subset of the official training data (10{,}000
images) is reserved for all model selection, and the official test split is
used only for the final comparison. Reported values are means and sample
standard deviations over five initializations, and paired comparisons use 95\%
Student-$t$ intervals. All trained models, including the linear classifier and
the references, use plain SGD; batch size is part of each method.
Hyperparameters and raw values appear in \cref{app:protocol}.

The references relax different constraints. \TeReLOffline{} uses the same
architecture and objective with subsequences of 128 and end-to-end
backpropagation. A layer-local network trained with a supervised contrastive
objective~\cite{khosla2020} (Layer-local SupCon) uses labels, minibatches, and
batch normalization at every layer. A supervised offline network is trained
end to end with labels and backpropagation. A random ReLU encoder of the same
shape, a linear probe on the input, and batch and incremental
SFA~\cite{wiskott2002,kompella2012} complete the comparison. Because the
methods differ in their data and optimization requirements
(\cref{tab:configuration}), the comparison measures representation quality at
each method's own budget rather than at equal compute.

Four ablations examine \TeReL{} itself: shuffling the stream, and setting the
temporal, variance, or covariance coefficient to zero.

\section{Results}
\label{sec:results}

\begin{figure}[H]
  \centering
  \includegraphics[width=\linewidth]{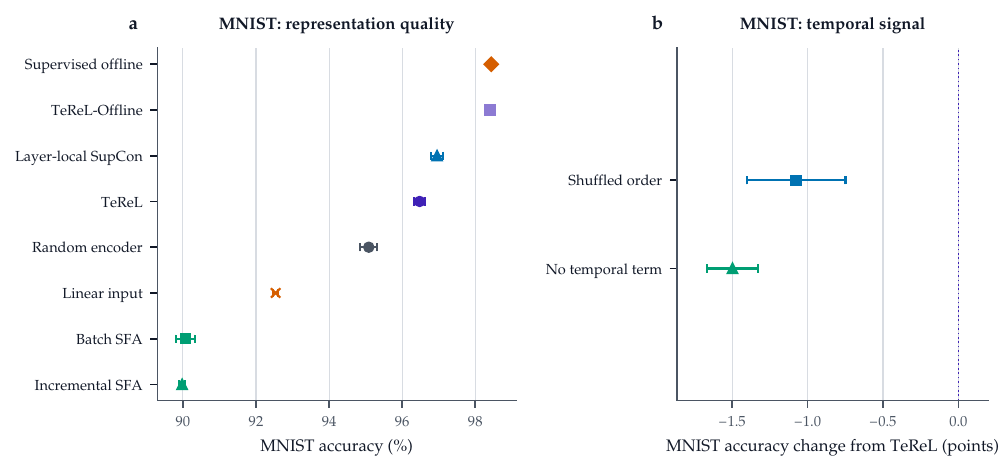}
  \caption{\textbf{Representation quality and temporal effects on MNIST.}
  \textbf{a}, Held-out linear-probe accuracy of each encoder (mean and sample
  standard deviation). \TeReL{}, trained with batch size one and local
  updates, lies above the random encoder and both SFA variants; \TeReLOffline{},
  the same objective without locality constraints, lies at the level of the
  supervised offline network. \textbf{b}, Paired accuracy change relative to
  \TeReL{} when the stream is shuffled or the temporal term is removed, with
  95\% Student-$t$ intervals; both lower accuracy.}
  \label{fig:performance}
\end{figure}

\subsection{Representation quality (H1)}

With batch size one and no gradient between layers, \TeReL{} reaches
$\FinalTeReLAccuracy\pm\FinalTeReLSD$\% held-out accuracy on MNIST
(\cref{fig:performance}a). The random encoder reaches
$95.09\pm0.23$\%, a paired gain of 1.39 percentage points $[1.04,1.74]$, and
the gain over a linear probe on the input is 3.94 points $[3.76,4.12]$. Batch
and incremental SFA reach $90.07\pm0.26$\% and $89.98\pm0.09$\%. Layer-local
SupCon, which uses labels and minibatches at every layer, reaches
$\FinalSupConAccuracy\pm\FinalSupConSD$\%, 0.48 points $[0.10,0.85]$ above
\TeReL{}. \TeReLOffline{} reaches
$\FinalOfflineAccuracy\pm\FinalOfflineSD$\%, on par with the fully supervised
offline network at $\FinalSupervisedAccuracy\pm\FinalSupervisedSD$\% (paired
difference $-0.03$ points $[-0.19,0.13]$). The objective itself therefore
supports supervised-level accuracy when
cross-layer and cross-time optimization are available; the gap to \TeReL{} is
the cost of the locality constraints.

\subsection{Temporal order and loss terms (H2)}

Removing the temporal order or any single loss term degrades the
representation (\cref{fig:performance}b). Shuffling the stream lowers accuracy
by 1.08 paired points $[0.75,1.40]$, and removing the temporal term lowers it by 1.50 points
$[1.33,1.67]$. Without the variance term, the representation collapses: the
median feature variance falls to about $1.4\times10^{-5}$ and accuracy to
$15.36\pm0.27$\%. Without the covariance term, training diverges for all five
initializations before a representation can be measured. These outcomes
match the roles derived in \cref{sec:method}: as in VICReg~\cite{bardes2022},
the invariance term alone admits a constant solution, which the variance term
prevents, and the covariance term keeps the multidimensional code stable.

\subsection{Representations}

\begin{figure}[H]
  \centering
  \includegraphics[width=0.88\linewidth]{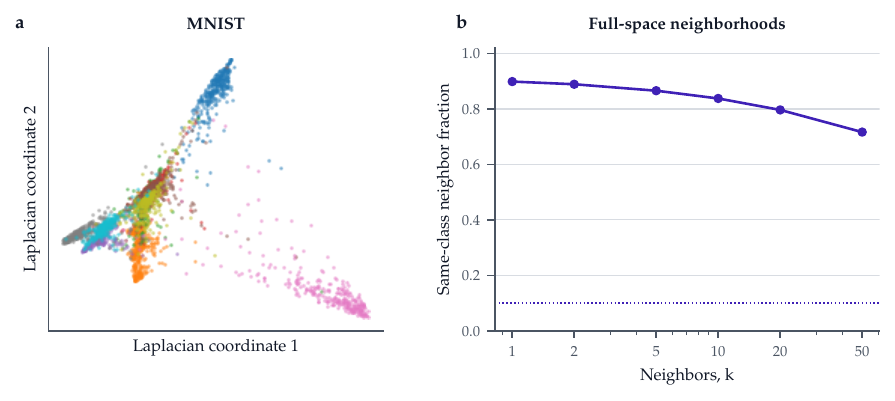}
  \caption{\textbf{Class structure in the MNIST \TeReL{} representation.}
  \textbf{a}, Laplacian embedding of validation features of both layers;
  colors denote digits and are used only for visualization. \textbf{b},
  Fraction of same-class neighbors in the complete 768-dimensional
  representation. The dotted line is the chance level. Most nearest neighbors
  share the digit, and the fraction stays far above chance up to 50
  neighbors.}
  \label{fig:geometry}
\end{figure}

The \TeReL{} representation is organized by class, which suits a linear
readout. The Laplacian embedding in \cref{fig:geometry}a shows
class-consistent branches rather than a single collapsed mass. Because a
two-dimensional projection can distort neighborhoods, \cref{fig:geometry}b
measures the same relation in the complete representation: the nearest
neighbor shares the digit for $89.9$\% of observations, and the fraction
remains $79.7$\% across 20 neighbors and $71.7$\% across 50, compared with a
$10.0$\% chance level.

\begin{figure}[!htbp]
  \centering
  \includegraphics[width=0.88\linewidth]{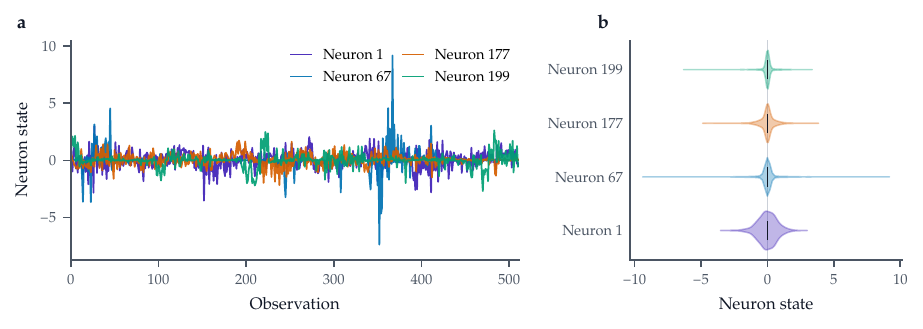}
  \caption{\textbf{Signed neuron state of \TeReL{}.} \textbf{a}, State
  trajectories of four randomly selected second-layer neurons over the same
  interval of the validation stream. \textbf{b}, Their state distributions
  over the complete validation stream; black bars mark medians. Each state
  takes both signs, so it drives both potentiation and depression of the
  feedforward synapses.}
  \label{fig:state}
\end{figure}

\Cref{fig:state} shows the neuron state $s_t$, the quantity that multiplies
presynaptic activity in \cref{eq:weight-update}. The selected neurons
alternate between positive and negative values, so the state drives both
potentiation and depression. Their scales differ: some neurons respond through
sparse excursions, whereas another carries a broader signed signal.

\section{Limitations and future work}

The experiments use MNIST streams whose chunks are ordered by class. Video and
sensor streams contain several time scales, and applying \TeReL{} to them is a
natural next step. A layer of width $D$ stores $D^2$ lateral parameters, which
exceed the feedforward parameters when a layer is wider than its input; sparse, low-rank, or
structured lateral matrices reduce this cost and suit constrained hardware.

\section{Conclusion}

Temporal Regularized Learning optimizes a VICReg-style objective one sample at
a time with updates that are local in space and time. Temporally adjacent
inputs supply the positives, detached running statistics and a lagged lateral
matrix supply the variance and covariance signals, and the gradient takes the
form of Hebbian feedforward and anti-Hebbian lateral updates driven by a signed
neuron state. Trained with batch size one, \TeReL{} learns representations that
improve on random features and classical SFA, and with its locality
constraints relaxed, the same objective matches supervised backpropagation.
\TeReL{} connects regularized self-supervised learning to Slow Feature Analysis
and trace learning, and because it learns from a stream with fixed-size state
at each neuron, it maps naturally to constrained hardware.

\section*{AI Use Statement}

An earlier version of \TeReL{} and of this manuscript predates the
generative-AI assistance. Generative AI tools were used primarily to update
the paper with new experiments, to find bugs, and for scripting. They also
assisted with analysis, figure creation, and editing. The author reviewed AI-assisted code, experimental
outputs, numerical claims, citations, figures, and text against saved
artifacts and executable code, and takes responsibility for the final content.

\bibliographystyle{unsrtnat}
\bibliography{references}

\appendix

\section{Evaluation protocol and raw values}
\label{app:protocol}

\Cref{tab:configuration} lists the reported configurations. All
hyperparameters are selected on the reserved validation split; the test split
is used only for the final comparison.

\begin{table}[ht]
\centering
\caption{Encoder configurations used in the reported comparison. A sample is
one observation presented to the encoder.}
\label{tab:configuration}
\small
\begin{tabularx}{\linewidth}{lYYYY}
\toprule
Method & Samples & Batch or subsequence & Optimizer & Normalization \\
\midrule
\TeReL{} & $100{,}000$ & $1$ & SGD & none \\
\TeReLOffline{} & $1{,}000{,}000$ & $128$ & SGD & none \\
Layer-local SupCon & $3{,}000{,}000$ & $256$ & SGD & BatchNorm \\
Supervised offline & $500{,}000$ & $64$ & SGD & none \\
Batch SFA & $50{,}000$ & complete fit & eigensolver & whitening \\
Incremental SFA & $50{,}000$ & $1$ & incremental & whitening \\
\bottomrule
\end{tabularx}
\end{table}

For \TeReL{}, the selected coefficients are
$\lambda_T=1$, $\lambda_V=3.001981$, $\lambda_C=0.606133$,
$\tau=1$, $\beta=0.99$, $\rho=0.995$, $\kappa=0.574950$,
$\eta=0.1$, and $\omega=0.591110$. The learning rate is $0.0138703$ with
weight decay $10^{-6}$. \TeReLOffline{} uses 20 epochs, learning rate
$0.4$, weight decay $10^{-6}$, variance coefficient
$2.529924$, covariance coefficient $0.0218471$, and variance target
$0.862789$. The supervised offline network uses learning rate $0.3$ and
Layer-local SupCon learning rate $0.3$. The linear probe is trained for 60
epochs with batch size 1024 on features standardized with training
statistics; its learning rate is selected per method on the validation split
from $\{0.01,0.03,0.1,0.3,1\}$ and is $0.3$ for \TeReL{} and its ablations.

\begin{table}[ht]
\centering
\caption{Held-out MNIST accuracy by seed. Values are
fractions; a dash denotes diverged training.}
\label{tab:raw}
\small
\begin{tabular}{lccccc}
\toprule
Method / seed & 1101 & 1202 & 1303 & 1404 & 1505 \\
\midrule
\TeReL{} & .9637 & .9647 & .9671 & .9633 & .9651 \\
\TeReLOffline{} & .9847 & .9837 & .9841 & .9836 & .9842 \\
Random encoder & .9475 & .9529 & .9495 & .9520 & .9524 \\
Input linear probe & .9254 & .9262 & .9257 & .9250 & .9246 \\
Supervised offline & .9836 & .9858 & .9840 & .9846 & .9838 \\
Layer-local SupCon & .9698 & .9713 & .9677 & .9711 & .9678 \\
Batch SFA & .9039 & .9010 & .8981 & .8981 & .9022 \\
Incremental SFA & .8984 & .8998 & .9001 & .9008 & .8998 \\
No temporal term & .9506 & .9503 & .9503 & .9478 & .9500 \\
No variance term & .1530 & .1537 & .1582 & .1515 & .1518 \\
No covariance term & -- & -- & -- & -- & -- \\
Shuffled order & .9559 & .9512 & .9535 & .9539 & .9556 \\
\bottomrule
\end{tabular}
\end{table}

\section{Locality and operation accounting}

For a layer of width $D$ with input width $d$, the feedforward multiplication
costs $Dd$ multiply--accumulates. Forming the target and running statistics is
linear in $D$. The lateral target and state correction each use a
$D\times D$ matrix--vector multiplication, and updating the lateral moment
uses a rank-one outer product. The persistent quantities are summarized in
\cref{tab:resources}; temporary activations are not persistent state.

\begin{table}[ht]
\centering
\caption{Persistent quantities for one \TeReL{} layer of width $D$ and input
width $d$.}
\label{tab:resources}
\small
\begin{tabular}{lll}
\toprule
Category & Quantity & Size \\
\midrule
Dynamic neuron state & $p,\mu,v$ & $3D$ \\
Auxiliary lateral parameters & $M$ & $D^2$ \\
Feedforward parameters & $W,q$ & $Dd+D$ \\
Optimizer state & none & $0$ \\
\bottomrule
\end{tabular}
\end{table}

\end{document}